%% file: emnlp2023.tex
\pdfoutput=1

\documentclass[11pt]{article}

\usepackage[]{EMNLP2023}

\usepackage{times}
\usepackage{latexsym}
\usepackage{subcaption}
\usepackage{arydshln}

\usepackage[T1]{fontenc}
\usepackage[]{xcolor}
\usepackage[utf8]{inputenc}

\usepackage{microtype}
\usepackage{multirow}

\usepackage{inconsolata}

\usepackage{graphicx}
\usepackage{breqn}
\usepackage{algorithm}
\usepackage[noend]{algpseudocode}
\usepackage[normalem]{ulem}
\usepackage{bm}
\usepackage{ragged2e}
\definecolor{esc}{RGB}{0, 153, 51}

\title{PiVe: Prompting with Iterative Verification \\ 
Improving Graph-based Generative Capability of LLMs}

\author{
  Jiuzhou Han$^{{\natural} }$\ \ \ \ \ 
  Nigel Collier$^{{\sharp}}$\ \ \ \ \ 
  Wray Buntine$^{{\flat}}$\ \ \ \ \ 
  Ehsan Shareghi$^{{\natural} }$\\
  $^{{\natural} }$~Department of Data Science \& AI, Monash University \\
  $^{{\flat}}$~College of Engineering and Computer Science, VinUniversity\\
  $^\sharp$~Language Technology Lab, University of Cambridge\\
  {jiuzhou.han@monash.edu}
}

\begin{document}
\maketitle
\begin{abstract}
Large language models (LLMs) have shown great abilities of solving various natural language tasks in different domains. Due to the training objective of LLMs and their pre-training data, LLMs are not very well equipped for tasks involving structured data generation. We propose a framework, Prompting with Iterative Verification (PiVe), to improve graph-based generative capability of LLMs. We show how a small language model could be trained to act as a verifier module for the output of an LLM~(i.e., ChatGPT, GPT-4), and to iteratively improve its performance via fine-grained corrective instructions. We also show how the verifier module could apply iterative corrections offline for a more cost-effective solution to the text-to-graph generation task. Experiments on three graph-based datasets show consistent improvement gained via PiVe. Additionally, we create GenWiki-HIQ and highlight that the verifier module can be used as a data augmentation tool to help improve the quality of automatically generated parallel text-graph datasets.\footnote{Our code and data are available at \url{https://github.com/Jiuzhouh/PiVe}.}
\end{abstract}

\section{Introduction}
Large language models (LLMs) like ChatGPT
and GPT-4 \citep{DBLP:journals/corr/abs-2303-08774} have been quite successful in solving different generative and reasoning tasks. The combination of their abilities in leveraging in-context learning as well as instruction following have unlocked new state-of-the-art results across the natural language processing (NLP) field. 
The existing LLMs are mostly pre-trained on a huge volume of unstructured data from the internet including books, articles, webtexts, repositories, Wikipedia, etc. Training on unstructured data naturally leads to relatively poor performance when dealing with tasks that demand organizing text into structured machine-readable format. 

A semantic graph, as a form of graph-structured data, stores information in a machine-accessible way \citep{DBLP:reference/fai/3}. Generating a semantic graph from text is  known as text-to-graph (T2G) generation and is previously attempted mostly by fine-tuning small language models \citep{DBLP:journals/corr/abs-2209-10754,DBLP:journals/corr/abs-2006-04702}. However, generating graph-structured data remains a challenge for LLMs even in the presence of reasonable number of few-shot examples. In fact, regardless of the number of few-shot examples or prompting style the outputs from LLMs (e.g., GPT-3.5) still contain errors and require correction (\S\ref{main_result}).  

\begin{figure}[t]
    \centering
    \includegraphics[scale=0.7]{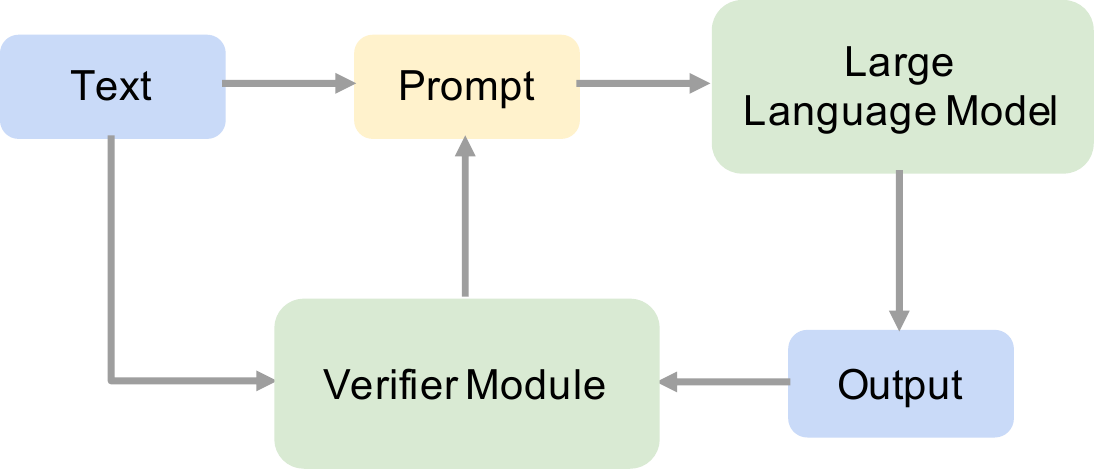}
    \caption{Framework of PiVe.}
    \label{fig:pive}
\end{figure}

In this paper, we focus on how to improve the graph-based generative capability of LLMs. To this end, we propose the Prompting through Iterative Verification (PiVe) framework shown in Figure~\ref{fig:pive}. Specially, PiVe involves leveraging an external verifier module (i.e., a much smaller LM) and incorporating the feedback from verifier module into the prompt. PiVe iteratively utilises the verifier module and refines the prompts, via corrective instructions, before sending them back into the LLM, leading to substantially improved quality of the generated semantic graphs.

In particular, to train the verifier modules, we start from a seed dataset of text and graph (\texttt{T,G}) pairs, and construct an arbitrarily large graph-perturbation dataset via a simple procedure which takes any graph $\texttt{G}$ from the seed set and perturbs it arbitrarily on its entities (\texttt{E}), relations (\texttt{R}), or triples~(\texttt{Tr}). The text and perturbed graph ($\bar{\texttt{G}}$), along with a corrective description to invert the applied perturbation (\texttt{IP}) form a verification dataset of (\texttt{T},$\bar{\texttt{G}}$,\texttt{IP}) triples  which serve as the training data for self-supervised learning of our verifier module. The verification dataset could be as large as desired~(i.e., for any seed dataset \texttt{D}, containing graphs of \texttt{|E|} entities, \texttt{|R|} relations, \texttt{|Tr|} triples, it could produce  $\mathcal{O}$(\texttt{|D|}$\times$\texttt{|E|}$\times$ \texttt{|R|}$\times$\texttt{|Tr|}) perturbations only by \emph{deleting}.\footnote{We also try other perturbation methods and show the results in Appendix \ref{Perturbation}.} We then devise fine-tuning and instruction-tuning to train domain-specific and unified verifiers, respectively. 

During the T2G generation via the LLM (e.g., in the zero-shot setting "\emph{Transform the text into a semantic graph: Text: ... Graph:}"), the verifier takes the text \texttt{T}, the output graph from the LLM, and sends a corrective signal to the LLM (e.g., "\emph{Transform the text into a semantic graph and add the given triples to the generated semantic graph: Text: ... Triples: ... Graph:}"). This process continues till the verifier module verifies the output as \emph{correct} and terminates. We refer to this as \emph{Iterative Prompting}. Additionally, there is another (more cost effective) mode to the verifier module, which starts by calling the LLM once at the start to get an initial graph, and then the rest of the corrective steps are all applied step-by-step and iteratively through the verifier offline. We refer to this as \emph{Iterative Offline Correction}.

Our extensive experiment results on three graph-based datasets demonstrate the effectiveness of the proposed PiVe framework in \emph{consistently} improving the quality of the LLM output via providing iterative corrective guidance by an average of 26\% across 3 datasets. We also create GenWiki-HIQ, a high-quality text-graph dataset and show how verifier module could be leveraged as a data augmentation technique to improve the quality of automatically constructed text-graph datasets.


\section{Basic Definitions}
\label{problem}
A semantic graph is a network that represents semantic relations between entities. Each semantic graph has its corresponding verbalisation, and can have different textual representations. A set of triples (i.e., \texttt{[subject, predicate, object]}) represents a semantic graph. Given a text, the task of text-to-graph generation is to query an LLM to generate a semantic graph of the text. The semantic graph should cover the information in the text as much as possible.

To prompt the LLM, we use few-shot 
by showing an example of T2G in the prompt to specify the expected format of the semantic graph (i.e., set of triples). We report experiments under various number of shots~(\S\ref{sec:moreshots}). The basic form of instruction we use in the prompt is "\textit{Transform the text into a semantic graph.}" followed by a text and a semantic graph pair as a demonstration. We also show results under different prompting strategies (\S\ref{sec:dprompts}). Different demonstrations are used for different datasets to adapt to the style of different datasets.
We show the used demonstrations in Appendix \ref{demo}.

\section{The PiVe Framework}
We first explain our training protocol for the verifier module (\S\ref{VM}), and then present our framework of iterative verification prompting~(\S\ref{IP}).
\subsection{Verifier Module}
\label{VM}
The quality of the generated semantic graph from LLM prompting could be quite poor. For instance the LLM often misses triples in the generated graph. In other words, some semantic relations between entities in the text are difficult to be captured for LLMs when they are generating a semantic graph. To detect the missing or incorrect parts of the generated semantic graph, we design a verifier module.

The verifier module is trained on a small pre-trained LM (\S\ref{model}). A typical graph-based dataset contains parallel text and semantic graph (\texttt{T,G}) pairs. For different graph-based datasets, we use their corresponding training data for the seed dataset to create data for the verifier module. In particular, for each text-graph pair in the original  dataset, we create one correct instance and one perturbed instance. We concatenate the text with graph using a separator token <S> and the target is to generate a specific output, denoted as \texttt{IP}, during training.
For correct instances, the \texttt{IP} is simply the word "Correct". For perturbed instances, we have two methods to create them: 
\begin{itemize}
    \item Random method: if the graph contains more than one triple, we randomly omit one triple from it and concatenate the text with perturbed graph using a separated token <S>. The target is to generate the missing part (e.g., triple \texttt{Tr}). 
    \item Heuristic method: Based on the observation that LLMs tend to miss the triples whose subject and object are not in the text, in addition to randomly omitting one triple from the graph we also omit the triple from the graph if subject and object of it are not in the text. 
\end{itemize}
The output to generate for perturbed examples is the missing triple, \texttt{Tr}. By utilising these two methods, we can create an arbitrarily large verification dataset to train a verifier module which will be used at inference time during prompting the LLM.
\subsection{Iterative Prompting}
\label{IP}
During the LLM prompting, the generated semantic graphs from LLMs is fed into the verifier module and the outputs from the verifier module is collected. If verifier generated "Correct" in its output, it means we do not need to make changes to the generated graph. Otherwise, the generated output from the verifier is added to the original prompt to create a new prompt. The new prompt is then used to query the LLM again. We repeat the whole process iteratively. The iteration process will stop when no missing triple is predicted or a maximum number of iterations is reached. 

\paragraph{New Prompt Design} As with the prompt used in the first iteration, we still provide an example in the new prompt for the subsequent iterations. The new prompt is "\textit{Transform the text into a semantic graph and also add the given triples to the generated semantic graph.}" In addition to the text, we also include some triples predicted by the verifier module which LLMs are likely to miss. This explicitly instructs the LLM to generate semantic graph and include the given triples. The given triple set contains the predicted missing triples from each iteration, which prevents the LLM from making the same mistakes as in previous iterations. See Appendix \ref{demo} for the used demonstrations.

\subsection{Iterative Offline Correction}
Similar to Iterative Prompting, the Offline Correction starts from the online LLM, but then continues with the step-by-step verification and correction steps offline. This approach is more cost effective as it relies only on one API call per instance (as opposed to several API calls of iterative prompting), however it is potentially weaker as it relies on the capability of the small verifier LM to both verify and apply the needed corrections. The offline correction stop under the same stopping criterion to Iterative Prompting.
\section{Experiments}
We describe the datasets and pre-processing method (\S\ref{DP}), introduce the models and implementation details (\S\ref{model}) and the evaluation metrics (\S\ref{metrics}). In Subsection \ref{main_result}, we describe the main result from PiVe, and compare the two modes of verifier: Iterative Prompting vs. Iterative Offline Correction (\S\ref{sec:correction}).  We then conduct various configurations of shots~(\S\ref{sec:moreshots}), and prompting~(\S\ref{sec:dprompts}). In Subsection \ref{dataaug}, we show how PiVe could be used for data augmentation of automatically generated graph-text datasets (e.g., GenWiki).

\subsection{Datasets and Preprocessing}
\label{DP}
We evaluate PiVe on three graph-based datasets, KELM \citep{DBLP:conf/naacl/AgarwalGSA21}, WebNLG+2020 \citep{DBLP:conf/inlg/GardentSNP17}, GenWiki \citep{DBLP:conf/coling/JinGQZ20}. 

\paragraph{KELM} is a large-scale synthetic corpus that consists of the English Wikidata KG and the corresponding natural text. It has $\sim$15M sentences synthetically generated using a fine-tuned T5 model. Each graph in KELM is a linearised KG containing a list of triples of the form \texttt{[subject, relation, object]}. If a triple has a sub-property, then it is quadruplet instead. We use a subset ($\sim$60K) of KELM which is named as KELM-sub. The creation of KELM-sub follows two criteria. We found that most graphs in KELM contain no more than six triples and only around 2,500 graphs contain more than six triples. Therefore, 1) we only consider the graphs with no more than six triples, and 2) we do not consider the graphs containing any triple with a sub-property. Based on these two criteria, for each size of graph (from one triple to six triples), we sampled equal number of (\texttt{T,G}) pairs. In total, the created KELM-sub contains 60,000/1,800/1,800 samples as train/validation/test set.

\paragraph{WebNLG+2020} contains a set of triples extracted from DBpedia \citep{DBLP:conf/semweb/AuerBKLCI07} in 16 distinct DBpedia categories and text description generated using diverse lexicalisation patterns. It contains $\sim$38K graphs and each graph has at most three different descriptions.

\paragraph{GenWiki} is a large-scale, general-domain dataset collected from general Wikipedia which contains 1.3 million non-parallel text and graphs with shared content. It has two versions: GenWiki\textsubscript{FULL} ($\sim$1.3M), and a fine version, GenWiki\textsubscript{FINE} ($\sim$750K), which adds constraints on the text and graphs to force them to contain highly overlapped entity sets. Both datasets are collected in a scalable and automatic way. GenWiki also has a human-annotated test set of 1,000 parallel text-graph pairs with high quality. Since both GenWiki\textsubscript{FULL} and GenWiki\textsubscript{FINE} are non-parallel text and graphs datasets, we cannot use it to train the verifier module. However, the relation types in GenWiki and WebNLG have some overlaps, so we use the verifier module trained on WebNLG+2020 and test it on GenWiki test set.


\begin{table}[t]
\centering
\scalebox{0.9}{
\begin{tabular}{lll}
\hline
\textbf{Dataset}  & \textbf{Train} & \textbf{Dev} \\ \hline
KELM-sub          & 110,000        & 3,300        \\
WebNLG \& GenWiki & 70,630         & 2,500      \\ 
\hline
\end{tabular}}
\caption{Statistics of the seed datasets for training the verifier modules on three datasets.}
\label{tab:statistics}
\vspace{-2mm}
\end{table}

We use the method described in Section \ref{VM} to create the data for training the verifier module. Table \ref{tab:statistics} shows the statistics of the created training data on these three seed datasets. 

\begin{table*}[ht]
\centering
\begin{tabular}{cccccc|cccc}
\hline
                          &             & \multicolumn{4}{c}{\textbf{Single Verifier Module}} & \multicolumn{4}{|c}{\textbf{Unified Verifier Module}} \\ 
                          &             & \textbf{T-F1↑}       & \textbf{G-F1↑}      & \textbf{G-BS↑}      & \textbf{GED↓}       & \textbf{T-F1↑}       & \textbf{G-F1↑}       & \textbf{G-BS↑}      & \textbf{GED↓}       \\ \hline
\multirow{4}{*}{KELM-sub} & Base & 13.50      & 4.89      & 83.92     & 13.20     & 13.50      & 4.89       & 83.92     & 13.20     \\
                          & Iteration 1 & 17.92      & 5.78      & 85.91     & 12.37     & 19.64      & 6.00       & 86.39     & 12.08     \\
                          & Iteration 2 & 19.46      & 6.44      & 86.57     & 12.08     & 22.11      & 6.44       & 87.31     & 11.68     \\
                          & Iteration 3 & 20.17      & 6.61      & 86.83     & 11.95     & 23.11      & 7.50       & 87.70     & 11.35     \\ \hline
\multirow{3}{*}{WebNLG}   & Base & 17.29      & 13.43     & 89.59     & 11.46     & 17.29      & 13.43      & 89.59     & 11.46     \\
                          & Iteration 1 & 18.32      & 14.00     & 89.74     & 11.23     & 18.22      & 13.83      & 89.67     & 11.23     \\
                          & Iteration 2 & 18.57      & 14.00     & 89.82     & 11.22     & 18.55      & 13.88      & 89.74     & 11.20     \\ \hline
\multirow{3}{*}{GenWiki}  & Base & 20.13      & 6.60      & 88.48     & 10.99     & 20.13      & 6.60       & 88.48     & 10.99     \\
                          & Iteration 1 & 20.54      & 6.80      & 88.70     & 10.87     & 20.88      & 6.70       & 88.66     & 10.90     \\
                          & Iteration 2 & 21.09      & 6.80      & 88.78     & 10.83     & 20.99      & 6.70       & 88.91     & 10.88  \\ \hline
\end{tabular}
\caption{Results of using PiVe on three datasets across all metrics. Single verifier module represents single dataset-specific verifier module trained on T5-Large and Unified verifier module is trained on Flan-T5-XXL using instruction-tuning.}
\label{tab:main_results}
\vspace{-1mm}
\end{table*}

\subsection{The LLM and Verifier Modules}
\label{model}
ChatGPT~(\texttt{gpt-3.5-turbo}) is used as our default LLM to perform the T2G task.\footnote{We also compare the graph-based generative capability between ChatGPT and GPT-3 in Appendix \ref{GPT-3} and report the fine-tuned results on small language model in Appendix~\ref{sec:fine-tune}.} We also experiment with GPT-4 in Subsection~\ref{sec:dprompts}. For verifier module, we use T5-Large \citep{DBLP:journals/jmlr/RaffelSRLNMZLL20}, and Flan-T5-XXL \citep{DBLP:journals/corr/abs-2210-11416} as the backbone models for dataset-specific verifier module, and unified verifier module, respectively. T5 models follow the encoder-decoder architecture and treat all NLP tasks as unified text-to-text transduction tasks. Flan-T5 is instruction-fine-tuned version of T5 which was trained on 1,836 NLP tasks initialized from fine-tuned T5 checkpoint. For T5-large, we fine-tune all parameters for separate verifier modules per each dataset. While for Flan-T5-XXL, we use LoRA \citep{DBLP:conf/iclr/HuSWALWWC22} as a parameter-efficient fine-tuning method, to train a unified verifier module which can follow the instruction. When using the unified verifier, we specify the dataset name in the instructions as datasets have different naming convention for relations.

The verifier is implemented using Pytorch \citep{DBLP:conf/nips/PaszkeGMLBCKLGA19} and Transformers \citep{DBLP:conf/emnlp/WolfDSCDMCRLFDS20}. For the training, we use Adam optimizer \citep{DBLP:journals/corr/KingmaB14}. Details about hyperparameter setting is provided in Appendix \ref{hyper}. For the implementation of parameter efficient training method used in Flan-T5-XXL, we use PEFT~\citep{peft} and 8-bit quantization technique \citep{DBLP:conf/iclr/DettmersLSZ22}. All training was done using a single A40 GPU with 48GB of RAM.

\subsection{Evaluation Metrics}
\label{metrics}
To evaluate the quality of the generated graphs given the ground-truth graphs, we use four automatic evaluation metrics: 

\textbf{Triple Match F1 (T-F1)} calculates F1 score based on the precision and recall between the triples in the generated graph and the ground-truth. We calculate the F1 scores for all test samples and compute the average F1 score as the final triple Match F1 score.

\textbf{Graph Match F1 (G-F1)} focuses on the entirety of the graph and evaluates how many graphs are exactly produced the same. For all test samples, we calculate the F1 score based on the precision and recall between all predicted graphs and all ground-truth graphs. This F1 score is the final Graph Match F1 score. Since graphs are represented in a linearised way, we could not simply use the string match method to check whether two graphs are the same. Instead, we first build directed graphs from linearised graphs using NetworkX \citep{los2008exploring}, then we consider the two graphs to be the same when all node and edge attributes match. 

\textbf{G-BERTScore (G-BS)}  is a semantic-level metric proposed by \citep{DBLP:conf/emnlp/SahaYBB21}, which extends the BERTScore \citep{DBLP:conf/iclr/ZhangKWWA20} for graph-matching. It takes graphs as a set of edges and solve a matching problem which finds the best alignment between the edges in predicted graph and those in ground-truth graph. Each edge is considered as a sentence and BERTScore is used to calculate the score between a pair of predicted and ground-truth edges. Based on the best alignment and the overall matching score, the computed F1 score is used as the final G-BERTScore. 

\textbf{Graph Edit Distance (GED)} \citep{DBLP:conf/icpram/Abu-AishehRRM15}  computes the distance between the predicted graph and the ground-truth graph. It measures how many edit operations (addition, deletion, and replacement of nodes and edges) are required for transforming the predicted graph to a graph isomorphic to the ground-truth graph. Lower GED between two graphs indicates the two graphs are more similar. In practice, the cost of each operation is set to be 1. For each sample, GED is normalized by a normalizing constant which is the upper bound of GED to make sure it is between 0 and 1. For demonstration, we multiply GED by 100 to show more decimals.


\begin{table*}[ht]
\centering
\scalebox{0.82}{
\begin{tabular}{ccccccc|ccccc}
\hline
                          &             & \multicolumn{5}{c}{\textbf{Iterative Prompting}}           & \multicolumn{5}{|c}{\textbf{Iterative Offline Correction}}      \\
                          &             &Time& \textbf{T-F1↑} & \textbf{G-F1↑} & \textbf{G-BS↑} & \textbf{GED↓} &  Time& \textbf{T-F1↑} & \textbf{G-F1↑} & \textbf{G-BS↑} & \textbf{GED↓} \\ \hline
\multirow{4}{*}{Single Verifier} & Base& 36.0 &13.50         & 4.89          & 83.92         & 13.20       & 36.0 & 13.50         & 4.89          & 83.92         & 13.20        \\
                          & Iteration 1 & +13.5  &17.92         & 5.78          & 85.91         & 12.37        & +4.5 & 17.76         & 5.83          & 86.42         & 12.37        \\
                          & Iteration 2 & +4.7 & 19.46         & 6.44          & 86.57         & 12.08        &+1.1& 18.51         & 6.11          & 86.91         & 12.19        \\
                          & Iteration 3 & +2.1 & 20.17         & 6.61          & 86.83         & 11.95        &+0.2& 18.55         & 6.17          & 86.94         & 12.18        \\ \hline
\multirow{4}{*}{Unified Verifier} & Base & 36.0 & 13.50         & 4.89          & 83.92         & 13.20        & 36.0 & 13.50         & 4.89          & 83.92         & 13.20        \\
                          & Iteration 1 & +118.9  & 19.64         & 6.00          & 86.39         & 12.08        & +105.0 & 16.99         & 5.67          & 87.08         & 12.95        \\
                          & Iteration 2 &+46.8& 22.11         & 6.44          & 87.31         & 11.68        &+40.6& 17.76         & 5.67          & 87.48         & 12.96        \\
                          & Iteration 3 &+21.1& 23.11         & 7.50          & 87.70         & 11.35        &+10.4& 17.85         & 5.67          & 87.52         & 12.96        \\ \hline     
\end{tabular}}
\caption{Comparison between Iterative Prompting and Iterative Offline Correction on KELM-sub dataset across all metrics using Single Verifier and Unified Verifier. Time denotes the total inference time in minutes.}
\label{tab:ip_vs_ma}
\end{table*}

\subsection{Results}
\label{main_result}

We report the evaluation results of using PiVe with ChatGPT on the test set of three datasets in Table \ref{tab:main_results}. All results presented under Base mean the direct output of the LLM without any verification. By utilising PiVe, on each dataset, we can see the consistent improvement of the quality of the generated graphs. For instance, in GenWiki which uses the same verifier module that was trained on the training data of WebNLG, the improvement of the scores over all metrics indicates the effectiveness of PiVe. Since the graphs are generated by the LLM through one-shot learning, G-F1 as the most strict metric, it is hard to get high G-F1 score (basically aiming for exact match without any minor deviation in wording, spelling, entities, or relations).

On WebNLG and GenWiki datasets, single verifier module performs slightly better than unified verifier module. While on KELM-sub dataset, unified module performs far better. We speculate this is due to the size of training data for KELM-sub verifier module being larger than that for WebNLG and GenWiki (as shown in Table \ref{tab:statistics}). Since unified verifier module combines the training data of different datasets, more training data leads to better performance for instruction-tuning. We conducted human evaluation which we include in Appendix \ref{human-eval} due to page limit.

\subsection{Iterative Prompting vs. Iterative Offline Correction} \label{sec:correction}
Instead of iteratively prompting the LLM, another way to utilise the results from verifier module is to append the predicted missing triples to the previously generated graph. The results of the comparison between iterative prompting and iterative offline correction using single verifier module and unified verifier module on KELM dataset is shown in Table \ref{tab:ip_vs_ma}. Iterative offline correction performs worse than iteratively prompting. This might be because iteratively prompting has the chance of doing self-correction. In each iteration, when we prompt the LLM, the generated graphs can probably correct the mistakes that were made in previous iteration. For example, in Figure~\ref{fig:example}, in Base the predicted relation regarding birth date is “birth year”, while the reference is “date of birth”. As the PiVe iteration continues, in Iteration 2, the relation “birth year” is regenerated as “date of birth” even though we didn't mention this in the prompt. Due to the page limit, we report the comparison results on WebNLG and GenWiki datasets in Appendix \ref{Additional}. Similarly, iterative prompting can achieve better results than iterative offline correction over all using different verifier modules.

\subsection{Impact of More Shots} \label{sec:moreshots} 
\begin{figure}[t]
    \centering
    \includegraphics[scale=0.97]{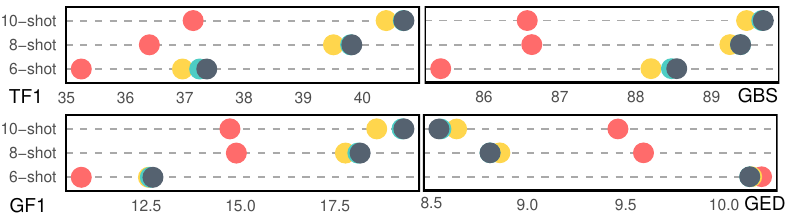}
    \caption{Results of various number of shots (k=6, 8, 10) on KELM-sub with Iterative Offline Correction. The colors represent \textcolor[HTML]{FF6B6B}{Base}, and corrective iterations \textcolor[HTML]{FFD64D}{1}, \textcolor[HTML]{4ECDC4}{2}, \textcolor[HTML]{556270}{3}.}
    \label{fig:more_shots}
    \vspace{-2mm}
\end{figure}

While in our main experiments, for cost reason, we used only one-shot demonstrations for the LLM prompting (i.e., GPT-3.5), we show that PiVe is effective in improving the results regardless of the underlying number of shots. Here we report the results of k-shot (k=6, 8, 10)
with the iterative offline correction (i.e., only using the LLM once to get the initial graph, while the correction steps are all applied step-by-step and offline). Figure~\ref{fig:more_shots} demonstrates the results 
on KELM-sub using unified verifier with iterative offline correction. The results show, as expected, that PiVe still provides consistent improvement even with the increase in the number of shots. Additionally, as the shots grow the improvement from PiVe also increases.

\subsection{Baselines on LLMs} \label{sec:dprompts}
To probe other prompting techniques as baselines of generating graphs from the LLM, we compare three diverse prompts. The first one we use is the default prompt used across our main experiments. This prompt is fairly direct and simple. Prompt 1:\texttt{ Transform the text into a semantic graph.} In the second prompt, we aim to instruct the LLM to generate larger graph with more triples. This is to increase the chance of LLM recovering more triples during the generation. Prompt 2:\texttt{ Transform the text into a semantic graph consisting of a set of triples. Generate as many triples as possible.} For the third prompt, inspired by Chain-of-thought~\citep{DBLP:conf/nips/Wei0SBIXCLZ22,DBLP:conf/nips/KojimaGRMI22} approach, we also ask the LLM to generate the semantic graph in two steps. Prompt 3:\texttt{ Transform the text into a semantic graph consisting of a set of triples. First produce all relations possible, then produce the graph.} 

\begin{figure}[t]
\centering
\begin{subfigure}[b]{0.25\textwidth}
\includegraphics[scale=0.63,trim={0 0cm 10cm 0},clip]{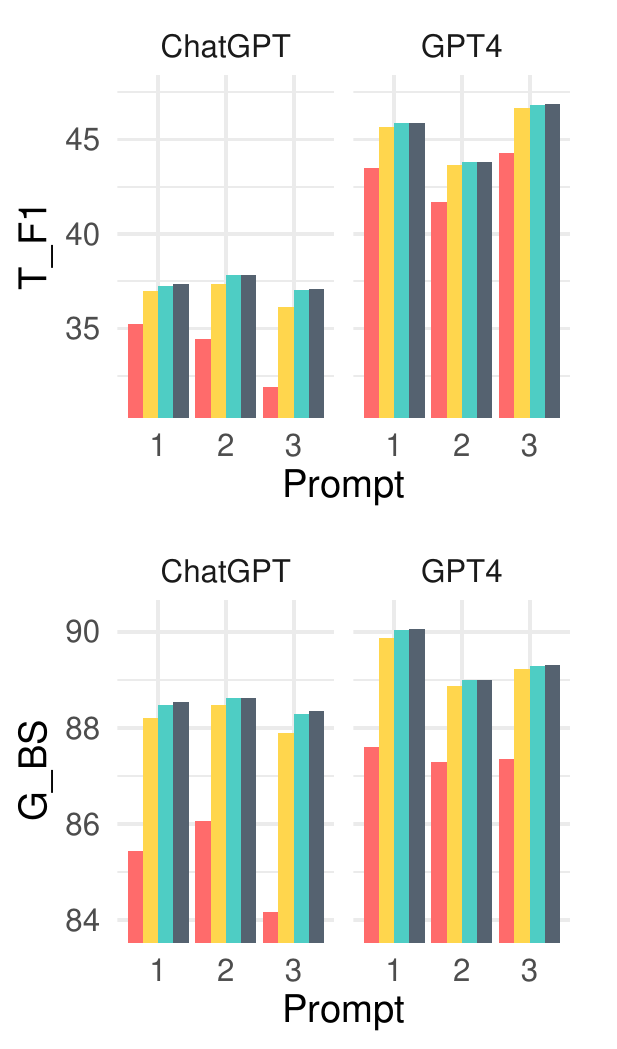}
\end{subfigure}%
\begin{subfigure}[b]{0.25\textwidth}
   \includegraphics[scale=0.63,trim={8cm 0 2cm 0},clip]{Figures/prompt.pdf}
\end{subfigure}
\caption{Results of using 3 diverse prompts with 6-shot on KELM-sub with Iterative Offline Correction on ChatGPT and GPT-4. The colors represent \textcolor[HTML]{FF6B6B}{Base}, and corrective iterations \textcolor[HTML]{FFD64D}{1}, \textcolor[HTML]{4ECDC4}{2}, \textcolor[HTML]{556270}{3}.}
\label{tab:prompts}
\vspace{-2mm}
\end{figure}

We conduct experiments on ChatGPT~(\texttt{gpt-3.5-turbo}) and {GPT-4}~(\texttt{gpt-4}) in 6-shot learning on KELM-sub, using unified verifier with iterative offline correction. The results are shown in  Figure~\ref{tab:prompts} (for detailed numbers see Table~\ref{tab:diverse_prompt_chatgpt} and Table~\ref{tab:diverse_prompt_gpt4} in Appendix). In general, as expected, GPT-4 performs far better than ChatGPT on the T2G task, but the effect of different prompts varies across these two models. Specifically, on ChatGPT, Prompt 2 achieves the best results while on GPT-4, Prompt 1 is outperforming the rest on most metrics. PiVe can consistently improve the results across all different settings, with the biggest jump in performance emerging in the first iteration, with slight improvements also observed between the second and third iterations of correction.

We also compare PiVe with the recent Self-Refine~\citep{DBLP:conf/nips/MadaanTGHGW0DPY23} method, which leverages LLM itself to provide feedbacks for self-refinement. Table~\ref{table:self_refine} shows the self-refine results on KELM-sub dataset. The results show that self-refine could not provide effective feedback, thus leading to the performance drop as the iteration goes. The performance gap is obvious comparing with our PiVe result. Since LLMs are not trained as rigorously on structured data compared to text, expecting them to provide meaningful feedbacks on their outputs is expected to fail.

\begin{table}[t]
\centering
\scalebox{0.9}{
\begin{tabular}{lcccc}
\hline
  & \textbf{T-F1↑} & \textbf{G-F1↑} & \textbf{G-BS↑} & \textbf{GED↓}  \\
\hline
Base & 13.50 & 4.89 & 83.92 & 13.20  \\
Iteration 1 & 14.54 & 3.00 & 84.85 & 13.90  \\
Iteration 2 & 14.33 & 2.44 & 84.57 & 14.43  \\
Iteration 3 & 13.79 & 2.28 & 84.20 & 15.18  \\
\hline
\end{tabular}}
\caption{Self-Refine~\citep{DBLP:conf/nips/MadaanTGHGW0DPY23} results on KELM-sub using Single Verifier.}
\label{table:self_refine}
\vspace{-2.5mm}
\end{table}

\subsection{Computational Cost and Trade-off}
\paragraph{Training and Inference}
The training and inference of both single verifiers and unified verifier are on a single A40 GPU. Each single verifier takes around 6 hours and the unified verifier takes around 40 hours to train. The computation cost for training of verifiers is a feasible one-off cost. Once the training is finished, the inference of the verification of each instance takes 0.15s for single verifier, and 3.5s for unified verifier. {See the Time column in Table~\ref{tab:ip_vs_ma}}. Different verifiers present performance-speed trade-offs and are significantly effective in augmenting the LLMs.

\paragraph{Stopping Criterion}
In theory, the verification module could run till no missing triple is predicted or a maximum number of iterations is reached. However, running more iterations increases the associated cost (i.e., OpenAI API charges). We set a maximum of 3 iterations.


\subsection{PiVe Examples}
\begin{figure*}[t]
    \centering
    \includegraphics[scale=0.75]{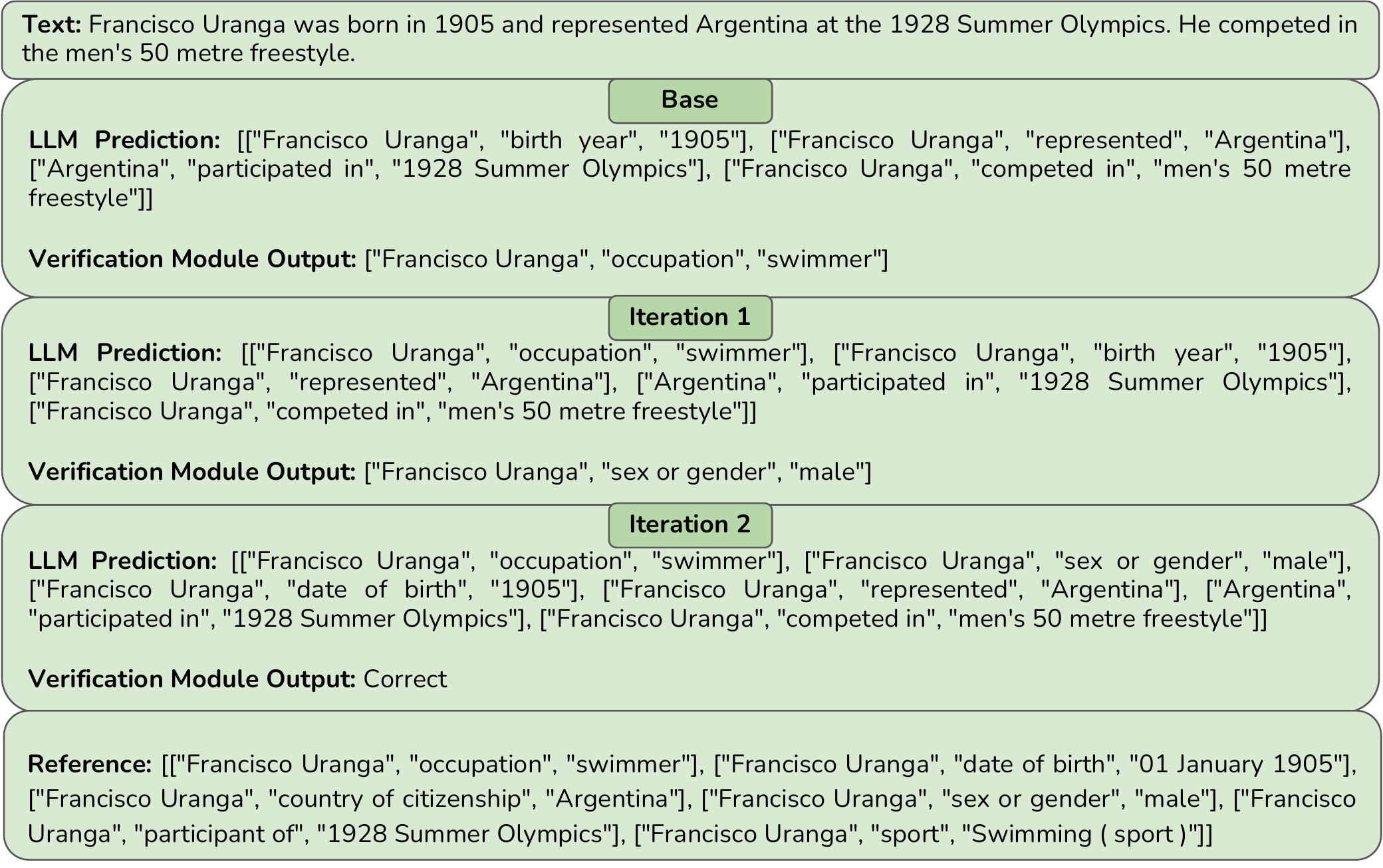}
    \caption{An example from KELM-sub test set using unified verifier module.}
    \label{fig:example}
\end{figure*}

In Figure \ref{fig:example}, we demonstrate an example from KELM-sub test set using unified verifier. In Base, based on the prediction from the LLM, the verifier module predicts the missing triple \texttt{[``Francisco Uranga'', ``occupation'', ``swimmer'']}. By suggesting this missing triple in the next iteration of prompt, the prediction from LLM includes it. Then, the verifier predicts the missing triple \texttt{[``Francisco Uranga'', ``sex or gender'', ``male'']}. In Iteration 2, both of these two missing triples are included in the prediction from LLM, and at this time, the verifier predicts ``Correct''. The prediction from Iteration 2 contains all information correctly in the reference. See another example in Appendix \ref{example2}. 

\section{GenWiki-HIQ}
\label{dataaug}
Training a good G2T or T2G model requires a large amount of high-quality parallel text and graph pairs or pre-training protocols to accommodate for lack of data~\cite{han-shareghi-2022-self}. However, creating the parallel data by human is a labour-intensive and time-consuming work. \citet{DBLP:conf/coling/JinGQZ20} proposed GenWiki, an automatically constructed large dataset containing non-parallel text and graphs with shared content. Although the text and graphs contain shared content, it can still only be used for unsupervised training due to the low entity and relation overlap between text and graph. Our verifier module can naturally serve as a data augmentation tool to improve the overlap between the text and graph of automatically constructed datasets. 


\paragraph{Iterative Augmentation.}
Based on GenWiki\textsubscript{FINE} ($\sim$750K), first we filter out the text that has little overlap with the graph. 
After filtering, we got around 110K text-graph pairs called GenWiki\textsubscript{FINE-f}. Then following the process described in Section \ref{VM}, we use the WebNLG verifier module and the iterative offline correction to improve the coverage and quality of GenWiki\textsubscript{FINE-f} and formed GenWiki-HIQ. The maximum number of iterations is four.


To evaluate the effectiveness of the verifier module as data augmentation tool, as well the quality of the generated graph, first we use Flan-T5-XL model to generate a description of each graph in zero-shot setting by using the prompt "\texttt{Transform the semantic graph into a description.}" for each iteration. Then we leverage automatic quality evaluation metrics to calculate the score between the generated description and the corresponding text. Ideally, the higher the similarity between the graph and the corresponding text, the higher the score of the generated description and corresponding text. We use four commonly used quality evaluation metrics which are BLEU \citep{DBLP:conf/acl/PapineniRWZ02}, METEOR \citep{DBLP:conf/acl/BanerjeeL05}, TER \citep{DBLP:conf/amta/SnoverDSMM06}, BERTScore.
\begin{table}[t]
\centering
\scalebox{0.85}{
\begin{tabular}{lcccc}
\hline
  & \texttt{BLEU↑} & \texttt{METEOR↑} & \texttt{TER↓} & \texttt{BERTScore↑}  \\
\hline
Base & 4.75 & 18.02 & 80.72 & 89.47  \\
Iteration 1 & 11.86 & 26.25 & 73.79 & 91.49  \\
Iteration 2 & 14.90 & 29.69 & 72.29 & 92.00  \\
Iteration 3 & 15.93 & 31.05 & 72.01 & 92.14  \\
\hline
\end{tabular}}
\caption{Results of iterative augmentation on filtered 110K text-graph pairs from GenWiki\textsubscript{FINE} across four metrics after each iteration. We take the text-graph pairs from Iteration 3 as the created GenWiki-HIQ dataset.}
\label{table:data_aug_result}
\vspace{-3mm}
\end{table}
\paragraph{Result.}
We used the dataset-specific verifier module to do the data augmentation. We conducted four iterations and the evaluation results are shown in Table \ref{table:data_aug_result}. The results in Base represent the scores over non-parallel graph-text pairs from GenWiki\textsubscript{FINE}, which have low overlap between graph and text. By using verifier module iteratively, we add more missing triples to the original graph, thus leading the higher quality scores. As the iteration progresses, fewer missing triples are added and we take the augmented graph-text pairs from the last iteration as the final created GenWiki-HIQ dataset. We also conducted G2T experiments in Appendix~\ref{G2T_genwiki} to further demonstrate the quality of GenWiki-HIQ. The G2T model trained on GenWiki-HIQ performs far better than the G2T model trained on GenWiki\textsubscript{FINE-f} on the human annotated GenWiki test set. This indicates that GenWiki-HIQ contains parallel text-graph pairs with high overlap.

\begin{figure}[t]
    \centering
    \includegraphics[scale=0.69]{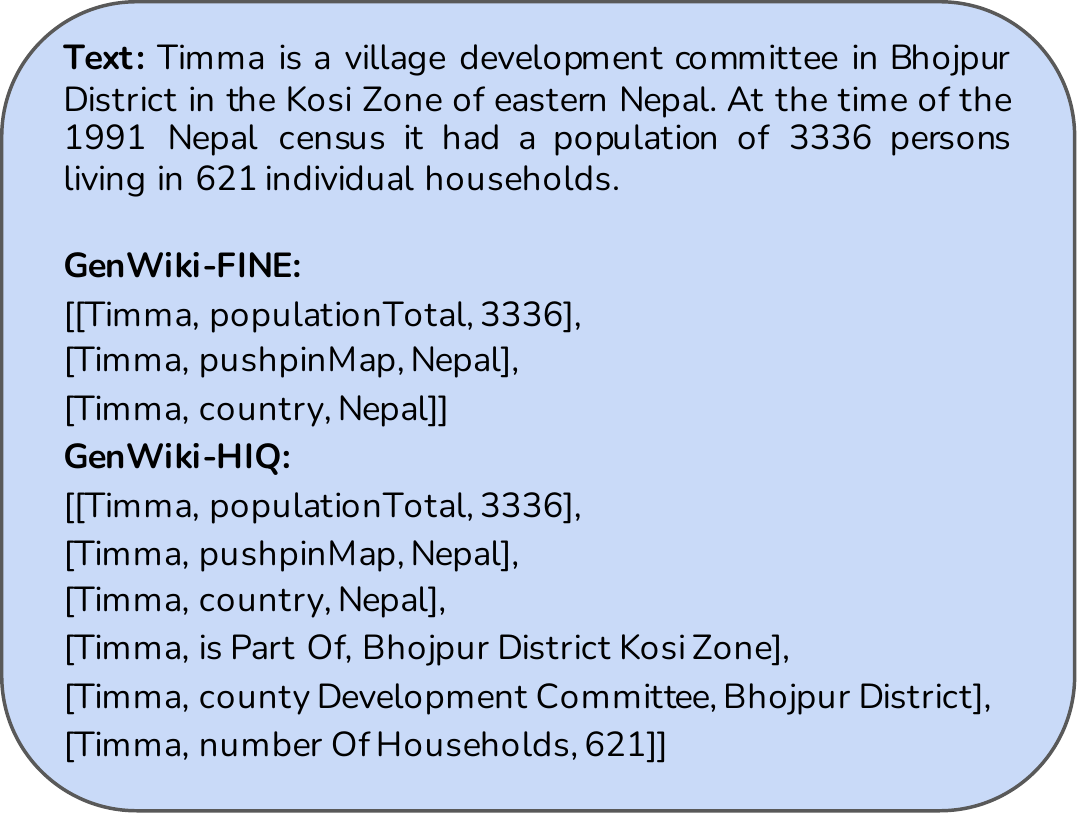}
    \caption{An example from GenWiki-HIQ compared to the original graph in GenWiki\textsubscript{FINE}.}
    \label{fig:genwiki_hiq}
    \vspace{-3mm}
\end{figure}

\paragraph{Qualitative Example.}
In Figure \ref{fig:genwiki_hiq}, we demonstrate an example from the created  GenWiki-HIQ dataset and the original graph in GenWiki\textsubscript{FINE}. After the data augmentation process, the graph in GenWiki-HIQ contains more information in text.

\subsection{G2T Results on GenWiki-HIQ}
\label{G2T_genwiki}
To further verify the quality of GenWiki-HIQ dataset, we use T5-large as the backbone model to train a G2T model, which generates the corresponding text based on the graph. Then we test it on the original GenWiki test set containing a 1,000 high-quality human annotated parallel text-graph pairs. As comparison, we also train another G2T model on GenWiki\textsubscript{FINE-f} which is the seed dataset of GenWiki-HIQ. 

The result is demonstrated in Table \ref{tab:genwiki_g2t}. On original Genwiki test set, the model trained on GenWiki-HIQ performs far better than the model trained on GenWiki\textsubscript{FINE-f} across all metrics. This indicates that GenWiki-HIQ contains parallel text-graph pairs with high overlap.

\begin{table}[t]
\centering
\scalebox{0.8}{
\begin{tabular}{lcccc}
\hline
  & \texttt{BLEU↑} & \texttt{METEOR↑} & \texttt{TER↓} & \texttt{BERTScore↑}  \\
\hline
GenWiki\textsubscript{FINE-f}  & 35.71   & 36.67    & 65.19   & 93.74  \\
GenWiki-HIQ & 48.17  & 42.03    & 41.94  &  95.44 \\
\hline
\end{tabular}}
\caption{Results of G2T generation on original GenWiki test set training on different datasets. GenWiki\textsubscript{FINE-f} contains the filtered 110K text-graph pairs from original GenWiki\textsubscript{FINE} as described in Section \ref{dataaug}. GenWiki-HIQ is the augmented dataset based on GenWiki\textsubscript{FINE-f}.}
\label{tab:genwiki_g2t}
\end{table}

\section{Background and Related Work}
\subsection{In-Context Learning}
With the scaling of model size and training corpus size \citep{DBLP:conf/nips/BrownMRSKDNSSAA20,DBLP:journals/corr/abs-2204-02311}, LLMs demonstrate new abilities of learning from a few demonstrations which contain some training examples \citep{DBLP:journals/corr/abs-2301-00234}. As a new paradigm, In-Context Learning does not require parameter updates and directly performs predictions on the pre-trained language models. The provided demonstration examples in the prompt follow the same format, which are usually written in natural language templates. By concatenating a query question with the demonstrations in the prompt, LLMs can learn from the given examples and make a prediction of the query question. Previous research~\citep{DBLP:conf/acl-deelio/LiuSZDCC22,DBLP:conf/acl/LuBM0S22} has shown that the number and order of the demonstrations can influence the In-Context Learning performance. These are further points of future investigation, which could potentially improve the initial graph produced by the LLM, which could further be corrected with the PiVe framework.

\subsection{Instruction-Tuning}
Instruction-Tuning \citep{DBLP:conf/acl/MishraKBH22,DBLP:conf/emnlp/WangMAKMNADASPK22,DBLP:journals/corr/abs-2301-13688} is a framework of doing multi-task learning, which enables the use of human-readable instructions to guide the prediction of LLMs. This novel training paradigm can improve the performance of the downstream tasks and also shows great generalisation ability on unseen tasks \citep{DBLP:journals/corr/abs-2210-11416,DBLP:conf/iclr/SanhWRBSACSRDBX22}. \citet{DBLP:journals/corr/abs-2304-08085} proposed a unified information extraction framework based on multi-task instruction-tuning. \citet{DBLP:journals/corr/abs-2304-14293} utilised instruction-tuning to perform controlled text generation following certain constraints. In our work, we use instruction-tuning to train a unified verifier module, which can follow the instruction to perform predictions on different datasets.

\subsection{Verifiers}
Leveraging small models could further improve the performance of LLMs. \citet{DBLP:journals/corr/abs-2110-14168} proposed to solve math word problem by utilising verifier. The verifier is used to judge the correctness of model-generated solutions. During test time, based on multiple candidate solutions generated, verifier calculates the correctness probability and the final answer will be selected by the verifier from the ranked list. \citet{welleck2023generating} proposed self-correction, an approach that trains a small model to iteratively apply self-correction. The idea of self-correction looks similar to our PiVe. While \citet{welleck2023generating} focuses on the design of a self-correcting language model, PiVe presents a very simple verifier module design and a simple data perturbation strategy to train such model. The ideas presented in our work are developed concurrently and independently.

\section{Conclusion}
We proposed PiVe, an iterative verification framework, to improve the graph-based generative capability of LLMs. We illustrated how a simple perturbation technique could be used to build data for training a verifier module which both verifies and corrects outputs from an LLM. We used different training strategies to build both dataset-specific verifiers with fine-tuning, and a unified verifier with instruction-tuning. Our verifier module could act both as an iterative prompting guide to improve outputs of an LLM, as well as an iterative offline correction system that starts from an LLM outputs but continuously improves it offline. The experimental results on three graph-based datasets demonstrates the effectiveness of PiVe. Furthermore, PiVe can also be used as a data augmentation technique to help improve the quality of automatically generated parallel text-graph datasets. By using verifier module, we created GenWiki-HIQ, a dataset containing 110K parallel text and graphs with high overlap for future research in text-graph domain.


\section*{Limitations}
Although the proposed framework is a straightforward and effective method of improving the generative capabilities of black box LLMs in graph generation, it still has some limitations. Firstly, PiVe is only designed for few-shot  prompting setting on LLMs, using an external verifier module to enhance their generative capabilities. The improvement is less significant when utilising PiVe on LMs that have been fine-tuned on the task data. Secondly, PiVe is not designed for free-form text generation tasks. Due to the unique aspect of graph, which has a specific structure, it allows for a much more fine-grained detection of errors and enables a richer corrective feedback. Translation between text and other similar modalities of data (e.g., table, SQL) can also effectively leverage our verification mechanism. Thirdly, in this work, we only focus on the triple missing mistake made by LLMs, so that the verifier module is not sensitive to the order of the head entity and tail entity. This means when the order of the head entity and tail entity in a triple of a generated graph from LLMs is incorrect, verifier module is not able to detect this type of mistake. It would be more effective if other error-detection heuristic methods are developed for creating the training dataset of the verifier. 

\section*{Ethics Statement}
Our work is built on top of existing pre-trained language models. Our goal was not to attend to alleviate the well-documented issues (e.g., privacy, undesired biases, etc) that such models embody. For this reason, we share the similar potential risks and concerns posed by these models.



\bibliography{custom}
\bibliographystyle{acl_natbib}

\appendix
\section*{Appendix}
\input{appendix}

\end{document}

%% file: Appendix.tex
\section{Hyperparameter Setting}
\label{hyper}

\begin{table}[t]
\centering
\scalebox{1.0}{
\begin{tabular}{lc}
\hline
\textbf{Hyperparameter}  & \textbf{Assignment} \\ \hline
Model   & T5-Large      \\
Epoch  & 5     \\ 
Batch Size  & 16     \\ 
Optimizer  & Adam     \\ 
Learning Rate & $2\times10^{-5}$     \\ 
Warm-up Step & 500    \\ 
Beam Size & 5    \\ 
\hline
\end{tabular}}
\caption{Hyperparameters of single verification module.}
\label{tab:hype1}
\end{table}

\begin{table}[t]
\centering
\scalebox{1.0}{
\begin{tabular}{lc}
\hline
\textbf{Hyperparameter}  & \textbf{Assignment} \\ \hline
Model  & FLAN-T5-XXL      \\
Epoch  & 2     \\ 
Batch Size  & 48     \\ 
Optimizer  & Adam     \\ 
Learning Rate & $3\times10^{-5}$     \\ 
Warm-up Step & 100     \\ 
Beam Size & 4    \\ 
\hline
\end{tabular}}
\caption{Hyperparameters of unified verification module.}
\label{tab:hype2}
\end{table}

\begin{table*}[t]
\centering
\scalebox{0.96}{
\begin{tabular}{cccccc|cccc}
\hline
                          &             & \multicolumn{4}{c}{\textbf{Iterative Prompting}}           & \multicolumn{4}{|c}{\textbf{Iterative Offline Correction}}      \\
                          &             & \textbf{T-F1↑} & \textbf{G-F1↑} & \textbf{G-BS↑} & \textbf{GED↓} & \textbf{T-F1↑} & \textbf{G-F1↑} & \textbf{G-BS↑} & \textbf{GED↓} \\ \hline
\multirow{3}{*}{Single Verifier}   & Base & 17.29         & 13.43         & 89.59         & 11.46        & 17.29         & 13.43         & 89.59         & 11.46        \\
                          & Iteration 1 & 18.32         & 14.00         & 89.74         & 11.23        & 18.03         & 13.55         & 89.16         & 11.52        \\
                          & Iteration 2 & 18.57         & 14.00         & 89.82         & 11.22        & 18.10         & 13.55         & 89.19         & 11.51        \\ \hline
\multirow{3}{*}{Unified Verifier}   & Base & 17.29         & 13.43         & 89.59         & 11.46        & 17.29         & 13.43         & 89.59         & 11.46        \\
                          & Iteration 1 & 18.22         & 13.83         & 89.67         & 11.23        & 18.02         & 13.49         & 89.21         & 11.61        \\
                          & Iteration 2 & 18.55         & 13.88         & 89.74         & 11.20        & 18.06         & 13.49         & 89.07         & 11.65        \\ \hline   
\end{tabular}}
\caption{Comparison between Iterative Prompting and Iterative Offline Correction on WebNLG dataset across all metrics using Single Verifier and Unified Verifier.}
\label{tab:ip_vs_ma2}
\end{table*}

\begin{table*}[t]
\centering
\scalebox{0.96}{
\begin{tabular}{cccccc|cccc}
\hline
                          &             & \multicolumn{4}{c}{\textbf{Iterative Prompting}}           & \multicolumn{4}{|c}{\textbf{Iterative Offline Correction}}      \\
                          &             & \textbf{T-F1↑} & \textbf{G-F1↑} & \textbf{G-BS↑} & \textbf{GED↓} & \textbf{T-F1↑} & \textbf{G-F1↑} & \textbf{G-BS↑} & \textbf{GED↓} \\ \hline
\multirow{3}{*}{Single Verifier}  & Base & 20.13         & 6.60          & 88.48         & 10.99        & 20.13         & 6.60          & 88.48         & 10.99        \\
                          & Iteration 1 & 20.54         & 6.80          & 88.70         & 10.87        & 20.24         & 6.70          & 89.00         & 10.95        \\
                          & Iteration 2 & 21.09         & 6.80          & 88.78         & 10.83        & 20.32         & 6.80          & 89.07         & 10.93 \\ \hline
\multirow{3}{*}{Unified Verifier}  & Base & 20.13         & 6.60          & 88.48         & 10.99        & 20.13         & 6.60          & 88.48         & 10.99        \\
                          & Iteration 1 & 20.88         & 6.70          & 88.66         & 10.90        & 20.37         & 6.60          & 89.08         & 10.96        \\
                          & Iteration 2 & 20.99         & 6.70          & 88.91         & 10.88        & 20.42         & 6.60          & 89.11         & 10.94   \\ \hline   
\end{tabular}}
\caption{Comparison between Iterative Prompting and Iterative Offline Correction on GenWiki dataset across all metrics using Single Verifier and Unified Verifier.}
\label{tab:ip_vs_ma3}
\end{table*}

\section{Additional Experiment Result}
\label{Additional}
Table \ref{tab:ip_vs_ma2} and Table \ref{tab:ip_vs_ma3} show the results of the comparison between iteratively prompt and iterative offline correction on WebNLG and GenWiki datasets.

\begin{table}[t]
\centering
\scalebox{0.8}{
\begin{tabular}{cccccc}
\hline
\multicolumn{1}{l}{}    &             & \textbf{T-F1↑} & \textbf{G-F1↑} & \textbf{G-BS↑} & \textbf{GED↓} \\ \hline
\multirow{4}{*}{Head}     & Base & 13.50          & 4.89           & 83.92          & 13.20         \\  
                          & Iteration 1 & 13.66          & 4.89           & 83.23          & 13.31       \\
                          & Iteration 2 & 13.65          & 4.89           & 81.99          & 13.31         \\
                          & Iteration 3 & 13.65          & 4.89           & 80.83          & 13.31         \\ \hline
\multirow{4}{*}{Relation} & Base & 13.50          & 4.89           & 83.92          & 13.20         \\  
                          & Iteration 1 & 15.09          & 4.94           & 83.68          & 12.97        \\
                          & Iteration 2 & 15.29          & 4.94           & 82.90          & 12.95      \\
                          & Iteration 3 & 15.33          & 4.94           & 82.09          & 12.96         \\ \hline
\multirow{4}{*}{Tail}     & Base & 13.50          & 4.89           & 83.92          & 13.20         \\ 
                          & Iteration 1 & 13.52          & 4.89           & 83.83          & 13.21         \\ 
                          & Iteration 2 & 13.51          & 4.89           & 83.74          & 13.22         \\ 
                          & Iteration 3 & 13.50          & 4.89           & 83.64          & 13.23         \\ \hline
\end{tabular}}
\caption{Results of doing different perturbations to the graph on KELM-sub to train a Single Verifier with Iterative Offline Correction.}
\label{tab:other_perturbations}
\end{table}

\begin{table}[t]
\centering
\scalebox{0.85}{
\begin{tabular}{lcccc}
\hline
  & \texttt{T-F1↑} & \texttt{G-F1↑} & \texttt{G-BS↑} & \texttt{GED↓}  \\
\hline
KELM-sub & 58.45 & 47.26 & 94.12 & 8.48  \\
WebNLG & 54.77 & 45.31 & 93.51 & 9.11  \\
GenWiki & 36.34 & 29.69 & 91.14 & 9.74  \\
\hline
\end{tabular}}
\caption{Fine-tuning results of text-to-graph generation on three datasets on T5-Large model.}
\label{table:finetune_result}
\end{table}

\begin{table*}[t]
\centering
\scalebox{0.96}{
\begin{tabular}{cccccc|cccc}
\hline
                          &             & \multicolumn{4}{c}{\textbf{Iterative Prompting}}           & \multicolumn{4}{|c}{\textbf{Iterative Offline Correction}}      \\
                          &             & \textbf{T-F1↑} & \textbf{G-F1↑} & \textbf{G-BS↑} & \textbf{GED↓} & \textbf{T-F1↑} & \textbf{G-F1↑} & \textbf{G-BS↑} & \textbf{GED↓} \\ \hline
\multirow{4}{*}{Single Verifier} & Base & 15.11          & 7.72          & 83.63          & 12.91        & 15.11          & 7.72          & 83.63          & 12.91        \\
                          & Iteration 1 & 19.55         & 8.78          & 85.90         & 11.98        & 18.97          & 8.72          & 86.47          & 12.10        \\
                          & Iteration 2 & 21.57         & 9.33          & 86.59         & 11.56        & 19.65          & 9.00          & 86.76          & 11.96        \\
                          & Iteration 3 & 22.49         & 9.89          & 87.10         & 11.37        & 19.69          & 9.00          & 86.77          & 11.95        \\ \hline
\multirow{4}{*}{Unified Verifier} & Base & 15.11          & 7.72          & 83.63          & 12.91        & 15.11          & 7.72          & 83.63          & 12.91        \\
                          & Iteration 1 & 21.40         & 8.78          & 86.43         & 11.69        & 20.46         & 8.78          & 87.18         & 11.90        \\
                          & Iteration 2 & 24.37         & 9.50          & 87.50         & 11.12        & 21.54         & 9.06          & 87.56         & 11.71        \\
                          & Iteration 3 & 26.06         & 10.22          & 87.99         & 10.83        & 21.57         & 9.06          & 87.57         & 11.71        \\ \hline     
\end{tabular}}
\caption{Results of using GPT-3-davinci as the backbone LLM on KELM-sub dataset over different settings.}
\label{tab:GPT-3}
\end{table*}

\begin{table}[t]
\centering
\scalebox{0.8}{
\begin{tabular}{cccccc}
\hline
\multicolumn{1}{l}{}    &             & \textbf{T-F1↑} & \textbf{G-F1↑} & \textbf{G-BS↑} & \textbf{GED↓} \\ \hline
\multirow{4}{*}{Prompt 1}     & Base & 35.25          & 10.78          & 85.43          & 10.19         \\  
                          & Iteration 1 & 36.96          & 12.56          & 88.20          & 10.14       \\
                          & Iteration 2 & 37.25          & 12.61          & 88.47          & 10.13         \\
                          & Iteration 3 & 37.37          & 12.68          & 88.54          & 10.13         \\ \hline
\multirow{4}{*}{Prompt 2}     & Base & 34.46          & 11.56           & 86.07          & 10.08         \\ 
                          & Iteration 1 & 37.38          & 14.22           & 88.48          & 9.35         \\ 
                          & Iteration 2 & 37.81         & 14.72          & 88.61         & 9.24         \\ 
                          & Iteration 3 & 37.85         & 14.89          & 88.62         & 9.23        \\ \hline
\multirow{4}{*}{Prompt 3}    & Base & 31.89          & 9.78          & 84.16          & 10.58         \\  
                          & Iteration 1 & 36.15          & 13.22          & 87.89          & 9.46       \\
                          & Iteration 2 & 37.03          & 13.94          & 88.29          & 9.23         \\
                          & Iteration 3 & 37.11          & 13.95          & 88.34          & 9.23         \\ \hline

\end{tabular}}
\caption{Results of using diverse prompts with 6-shot learning on KELM-sub with Iterative Offline Correction on ChatGPT.}
\label{tab:diverse_prompt_chatgpt}
\vspace{-3.5mm}
\end{table}

\begin{table}[t]
\centering
\scalebox{0.8}{
\begin{tabular}{cccccc}
\hline
\multicolumn{1}{l}{}    &             & \textbf{T-F1↑} & \textbf{G-F1↑} & \textbf{G-BS↑} & \textbf{GED↓} \\ \hline
\multirow{4}{*}{Prompt 1}     & Base & 43.46          & 26.50          & 87.60          & 7.97         \\  
                          & Iteration 1 & 45.68          & 32.50          & 89.86          & 7.39       \\
                          & Iteration 2 & 45.87          & 33.06          & 90.04          & 7.32         \\
                          & Iteration 3 & 45.87          & 33.06          & 90.05          & 7.31         \\ \hline
\multirow{4}{*}{Prompt 2}     & Base & 41.67          & 22.67           & 87.28          & 8.47         \\ 
                          & Iteration 1 & 43.64          & 26.61           & 88.87          & 7.98         \\ 
                          & Iteration 2 & 43.79         & 27.22          & 88.99         & 7.91         \\ 
                          & Iteration 3 & 43.79         & 27.28          & 89.00         & 7.91        \\ \hline
\multirow{4}{*}{Prompt 3}     & Base & 44.30          & 23.89           & 87.36          & 8.11         \\ 
                          & Iteration 1 & 46.65          & 29.61           & 89.22          & 7.49        \\ 
                          & Iteration 2 & 46.84          & 30.06           & 89.28          & 7.45         \\ 
                          & Iteration 3 & 46.85          & 30.08           & 89.30          & 7.45         \\ \hline

\end{tabular}}
\caption{Results of using diverse prompts with 6-shot learning on KELM-sub with Iterative Offline Correction on GPT-4.}
\label{tab:diverse_prompt_gpt4}
\end{table}

\section{Effect of Perturbation Method}
\label{Perturbation}
As described in Section \ref{VM}, we perturb the graph by omitting one triple when building the verifier module of PiVe. In addition, we also investigated other perturbation methods to train a verifier module, such as perturbing the head entity, relation and tail entity. To be specific, for head entity perturbation, if the graph contains more than one triple, we randomly choose one triple and replace the head entity with a different head entity from other triples of the same graph. Likewise, we replace the relation and tail entity for relation perturbation and tail entity perturbation, respectively. The target is to predict the original triple. Then we train different verifier modules using these three perturbation methods on KELM-sub.     The results of doing different perturbations using Iterative Offline Correction is shown in Table \ref{tab:other_perturbations}. 

Comparing with the result of omitting triple perturbation method shown in Table \ref{tab:ip_vs_ma} using Single Verifier with Iterative Offline Correction, these three perturbation methods have varying effects. While the relational perturbation works in terms of T-F1, with more iterations, the G-BS score generally goes down for all these perturbations. This indicates the verifier module could potentially inject wrong corrections if not trained with the proper perturbation mechanism. We speculate the reason is because LLMs are less likely to make mistakes at entity level, so these perturbation methods are not useful for training a verifier module. This also indicates when building a verifier module, choosing reasonable perturbation methods is significant and necessary.

\section{ChatGPT vs GPT-3}\label{GPT-3}
To further highlight the generalisation ability of PiVe, in addition to ChatGPT, we also experiment with GPT-3 (\texttt{text-davinci-003}) as the backbone LLM to perform the T2G task. We perform experiments on KELM-sub dataset using iterative prompting and iterative offline correction with different verifiers. The results are shown in Table~\ref{tab:GPT-3}. Compared with the results of using ChatGPT (shown in Table \ref{tab:ip_vs_ma}), GPT-3 has a better graph-based generative capability. Nonetheless, PiVe can still consistently further improve its results over all settings. Using iterative prompting with the unified verifier can achieve the best result on KELM-sub.


\section{Comparison with Fine-tuned Baselines}\label{sec:fine-tune}
While our work focuses on the fundamental question of "How can we improve the generative capabilities of black box LLMs in graph generation?", for completeness we also provide results of fine-tuned T5~\cite{DBLP:journals/jmlr/RaffelSRLNMZLL20} in Table~\ref{table:finetune_result}. As expected, fine-tuning on large amount of data surpasses few-shot prompting. This underscores the struggle LLMs face in transduction problems, and the need for additional mechanisms (like PiVe) to help LLM improvement. 

\section{Human Evaluation}\label{human-eval}
We conducted a human evaluation on 105 randomly sampled instances from three datasets (KELM-sub, WebNLG, GenWiki). Specifically, for each dataset, first we took the test set outputs from the first iteration and the last iteration, then we randomly sampled 35 instances from those with different outputs. The output from the first iteration is the original ChatGPT output without using PiVe, and the output from the last iteration is the result after using PiVe. For the evaluation process we recruited three annotators (1 PhD graduate and 2 PhD students in Computer Science and NLP) to select, for a given text and two graph outputs, which graph matches the text better. Each annotator should only choose one graph per each instance and evaluate all 105 instances. 

After annotation, we took majority voting over the result of each instance, then calculated the number of wins for ChatGPT with or without PiVe. The results are shown in Table \ref{tab:human_eval_result}. From the results, we can see ChatGPT with PiVe wins on 85 out of 105 samples and the total winning rate is over 80\%. This indicates the PiVe can effectively improve the graph-based generative capability of LLMs.

\begin{figure*}[t]
    \centering
    \includegraphics[scale=0.75]{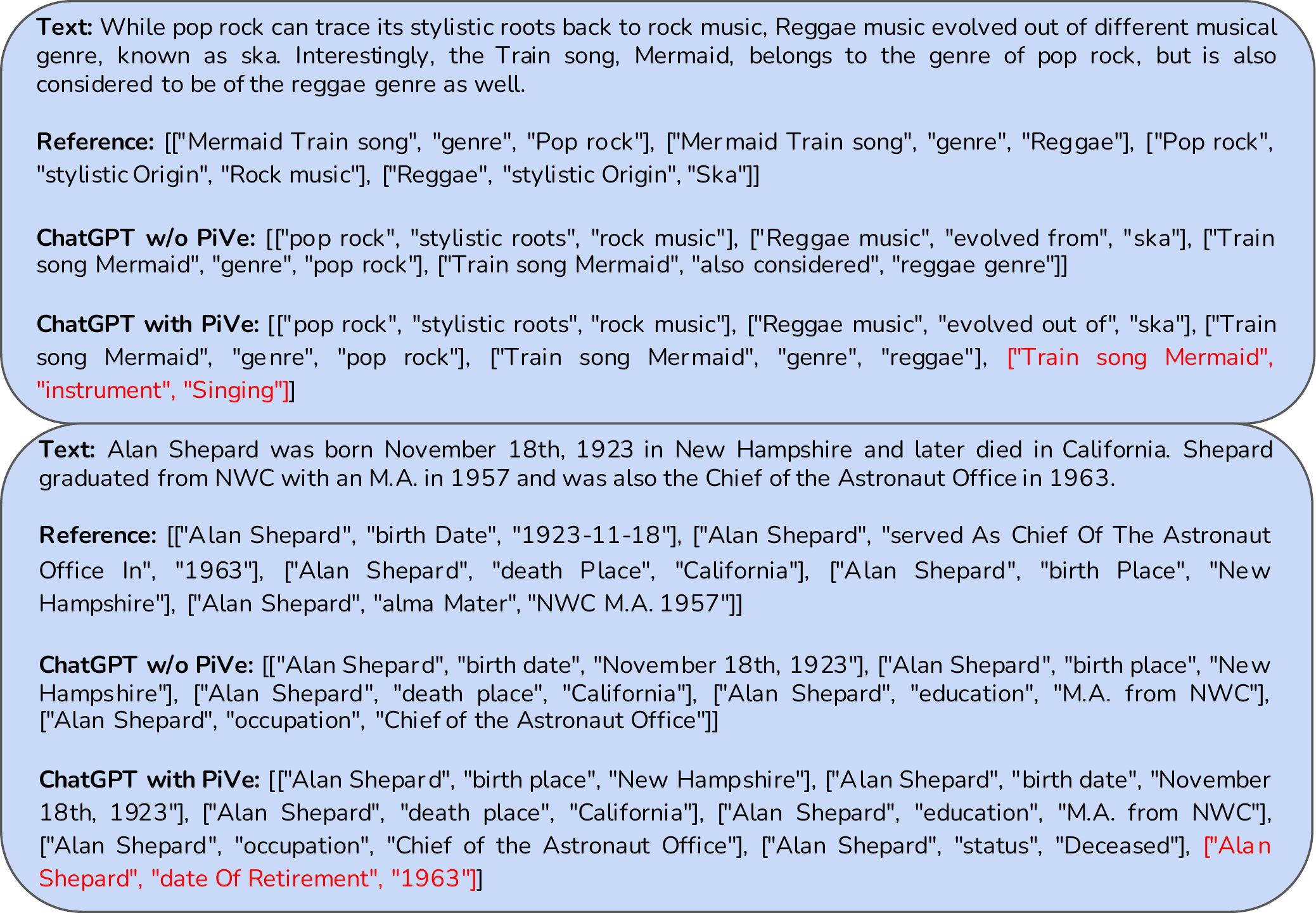}
    \caption{Two examples of PiVe making two types of mistakes: redundancy and inaccuracy.}
    \label{fig:errors}
\end{figure*}

For the cases that PiVe failed to improve, we did error analysis and found that there were mainly two types of mistakes that PiVe made: redundancy and inaccuracy. In Figure \ref{fig:errors}, we demonstrate two examples containing these two types of mistakes shown in red text. In the first example, the triple \texttt{[``Train song Mermaid'', ``instrument'',``Singing'']} predicted by PiVe is redundant. In the second example, the relation \texttt{``date Of Retirement''} in the triple \texttt{[``Alan Shepard'',``date Of Retirement",``1963'']} is inaccurate. We speculate these behaviours were caused due to the presence of many similar texts with similar graphs in the training data. During training, PiVe learned the potential connections between these similar graphs, thus leading to redundant and inaccurate triples at prediction.

\section{Demonstrations in Prompt}\label{demo}
\begin{figure*}[t]
    \centering
    \includegraphics[scale=0.75]{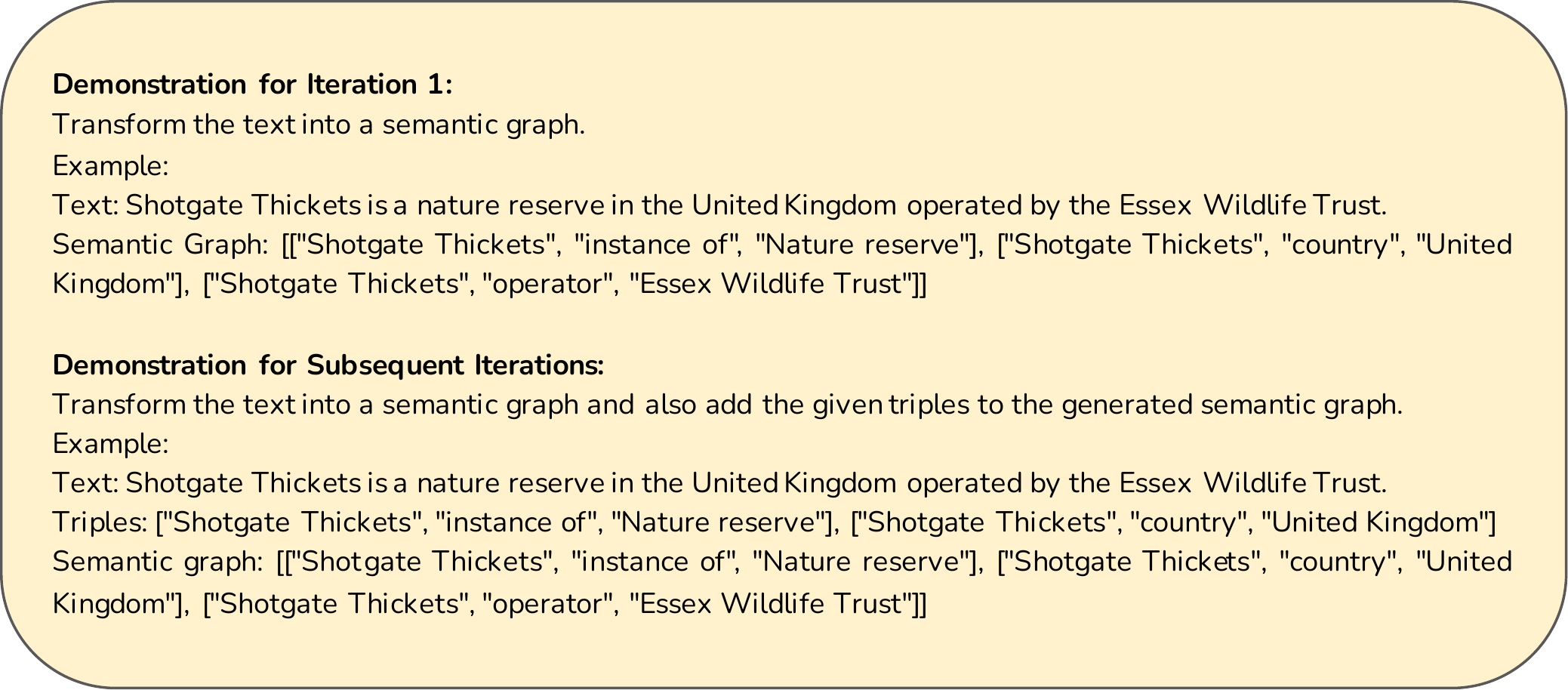}
    \caption{The demonstrations used in prompt for KELM-sub.}
    \label{fig:kelm_demo}
\end{figure*}

\begin{figure*}[t]
    \centering
    \includegraphics[scale=0.75]{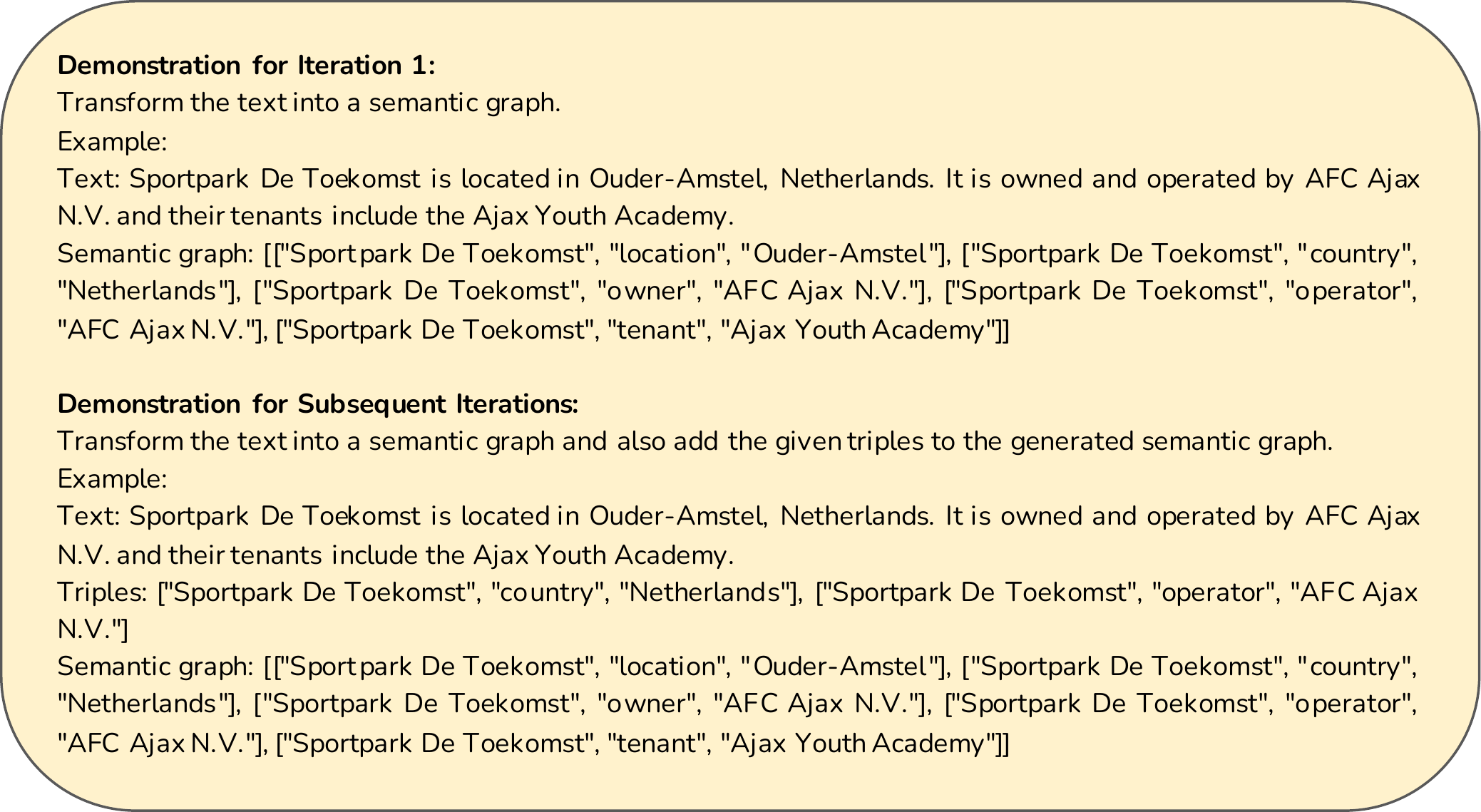}
    \caption{The demonstrations used in prompt for WebNLG and GenWiki.}
    \label{fig:webnlg_demo}
\end{figure*}

Figure \ref{fig:kelm_demo} shows the demonstrations used for KELM-sub and Figure \ref{fig:webnlg_demo} shows the demonstrations used for WebNLG and GenWiki. In Iteration 1, we use the demonstration that does not contain the missing triples. For subsequent iterations, we include the missing triples in the demonstration.

\section{PiVe Examples}\label{example2}

\begin{figure*}[t]
    \centering
    \includegraphics[scale=0.75]{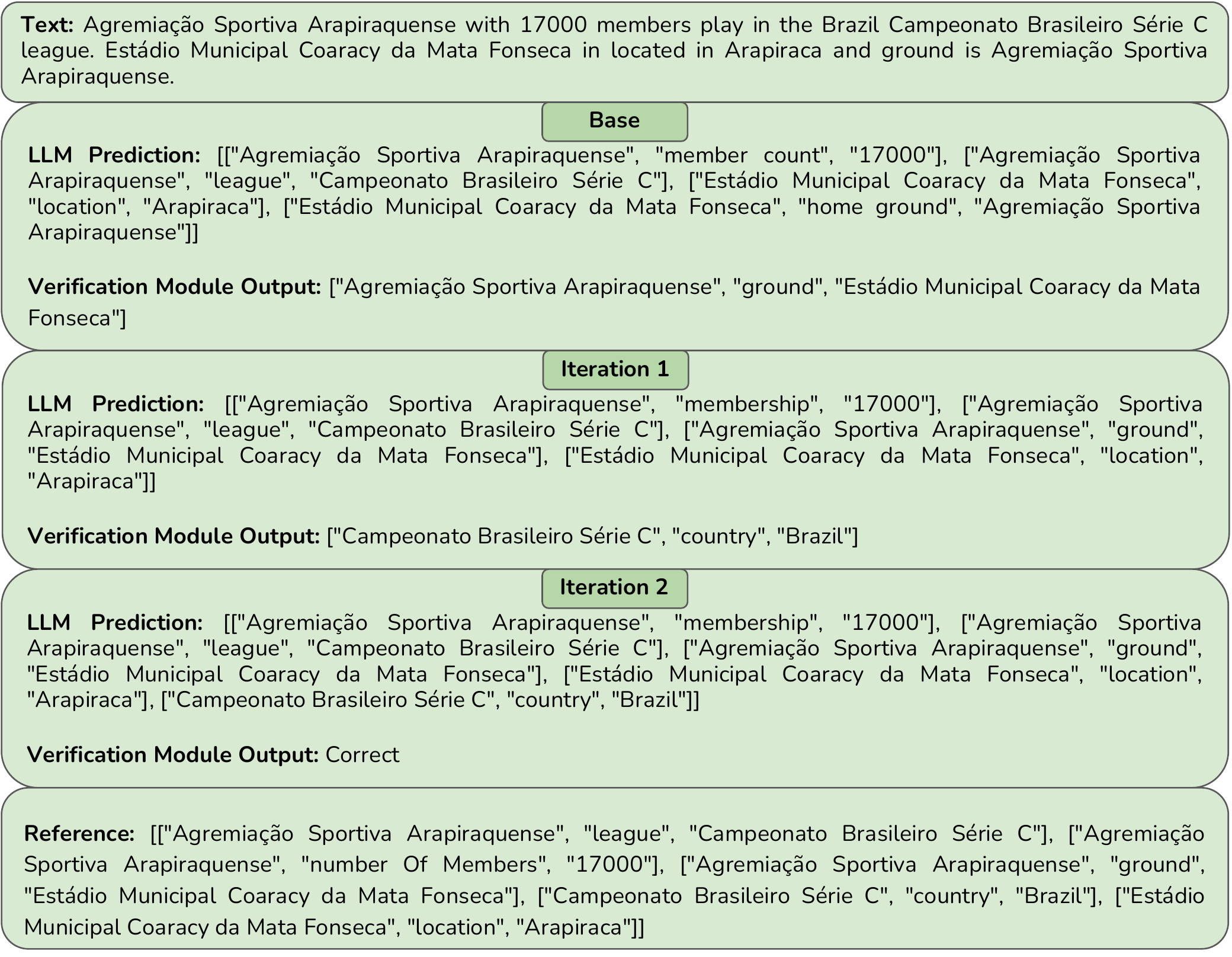}
    \caption{An example from WebNLG test set using single verification module.}
    \label{fig:example2}
\end{figure*}

Figure \ref{fig:example2} illustrates another example of PiVe from WebNLG test set using single verification module. In Base, the verification module predict the missing triple \texttt{[``Agremiação Sportiva Arapiraquense'', ``ground'', ``Estádio Municipal Coaracy da Mata Fonseca'']}, even though there is a similar triple but containing mistakes in the prediction from the LLM. In Iteration 1, the LLM corrects the mistakes in the previous iteration, and also includes the predicted missing triple. Based on the prediction from the LLM, the verification module predict the missing triple \texttt{[``Campeonato Brasileiro Série C'', ``country'', ``Brazil'']}. In Iteration 2, the verification module predict ``Correct'' to the final prediction from the LLM. After three iterations using PiVe, the predicted graph contains all information in the reference.


\begin{table}[t]
\centering
\scalebox{0.9}{
\begin{tabular}{lcc}
\hline
\textbf{Dataset}  & \textbf{\# with PiVe wins} & \textbf{\# w/o PiVe wins} \\ \hline
KELM-sub & 31        & 4        \\
WebNLG & 28         & 7      \\ 
GenWiki & 26         & 9      \\ \hline
Total & 85         & 20      \\
\hline
\end{tabular}}
\caption{Human evaluation results on 105 samples from three datasets using ChatGPT with or w/o PiVe.}
\label{tab:human_eval_result}
\vspace{-2.5mm}
\end{table}

